\documentclass[sigconf,nonacm]{acmart}

\AtBeginDocument{%
  }

\usepackage{multirow}
\usepackage{colortbl}
\usepackage{enumitem}

\begin{document}

\title[AMRS: A Rollout-Based World Model for Offline Preference Optimization]{Affective Music Recommendation: A Rollout-Based World Model for Offline Preference Optimization}

\author{Audrey Chan}
\affiliation{%
  \institution{LUCID Inc.}
  \city{Toronto}
  \country{Canada}}
\email{audrey@thelucidproject.ca}

\author{Aaron Labb\'{e}}
\affiliation{%
  \institution{LUCID Inc.}
  \city{Montr\'{e}al}
  \country{Canada}}
\email{aaron@thelucidproject.ca}

\author{Jacob Lavoie}
\affiliation{%
  \institution{Mila --- Qu\'{e}bec AI Institute}
  \city{Montr\'{e}al}
  \country{Canada}}
\email{jacob.lavoie@mila.quebec}

\author{Jordan Bannister}
\affiliation{%
  \institution{Mila --- Qu\'{e}bec AI Institute}
  \city{Montr\'{e}al}
  \country{Canada}}
\email{jordan.bannister@mila.quebec}

\author{Ars\`{e}ne Fansi Tchango}
\affiliation{%
  \institution{Mila --- Qu\'{e}bec AI Institute}
  \city{Montr\'{e}al}
  \country{Canada}}
\email{arsene.fansi.tchango@mila.quebec}

\author{Guillaume Lajoie}
\affiliation{%
  \institution{Mila --- Qu\'{e}bec AI Institute}
  \city{Montr\'{e}al}
  \country{Canada}}
\email{guillaume.lajoie@mila.quebec}

\author{Laurent Charlin}
\affiliation{%
  \institution{Mila --- Qu\'{e}bec AI Institute}
  \city{Montr\'{e}al}
  \country{Canada}}
\email{lcharlin@mila.quebec}

\renewcommand{\shortauthors}{Chan et al.}

\begin{abstract}
Functional music applications, from consumer focus and sleep aids to clinical interventions, share a distinctive recommendation problem: success is defined by the listener's \emph{affective state}, but online experimentation on emotion is ethically constrained, particularly for clinical populations who cannot reliably skip a song or report distress. We describe AMRS, the Affective Music Recommendation System deployed on LUCID's health-and-wellness platforms, which serve clinical users (primarily older adults with neurocognitive conditions) and consumer-wellness users across energize, focus, calm, and sleep modes. AMRS is built around a \emph{rollout-based world model}: a causal transformer trained on logged listening data to jointly predict engagement, binary rating, and self-reported valence and arousal. The world model serves both as an in-silico simulator for offline policy training and as a stress-testing tool before deployment. A recommender policy initialized by behaviour cloning is fine-tuned offline with Direct Preference Optimization (DPO) against a configurable multi-objective utility function. Under a strict cold-start protocol, the world model predicts both behavioural and affective signals with usable fidelity; DPO improves predicted valence and arousal over the cloned baseline while maintaining a similar diversity profile and avoiding the distributional collapse produced by greedy optimization. We position the work as an early deployed validation of a methodology for affective recommendation when online experimentation is ethically untenable.
\end{abstract}



\maketitle

\section{Introduction}
\label{sec:intro}

LUCID\footnote{\url{https://www.lucidtherapeutics.com}} operates a health-and-wellness music platform spanning two applications on shared infrastructure: a \emph{clinical} application for populations with mental health and neurocognitive conditions (primarily older adults, many living with dementia or mild cognitive impairment), and a \emph{consumer} wellness application for functional listening in four modes (energize, focus, calm, and sleep). Each session delivers a curated sequence of songs chosen by a fixed production policy. The platform logs engagement (the fraction of a song played) and a binary like/dislike rating at the song level, and a pre- and post-session pair of self-reported valence and arousal on the circumplex~\cite{russell1980circumplex} at the session level.

The success metric in both applications is not engagement: it is the listener's \emph{affective state}~\cite{russo2023developing}. A song that attracts attention does not necessarily produce the intended emotional trajectory. Building a recommender for this setting introduces three constraints that together break with the standard approach.

\paragraph{Two qualitatively different feedback signals.} Behavioural feedback (engagement, rating) is observed at the song level and is relatively plentiful. Affective feedback (valence, arousal) is observed at the session level (roughly one pair per multi-song session). Although functional music platforms such as Headspace, Calm, and Spotify's playlists reach millions of listeners, to our knowledge none collect self-reported affective feedback at comparable scale, leaving our most relevant signal the most sparsely supervised.

\paragraph{Ethical constraints on online experimentation.} When the optimization target is a listener's affective state, online experimentation is qualitatively different from standard A/B testing over engagement. Exploratory recommendations that shift valence or arousal in an unvalidated way carry real risk for vulnerable populations: people living with dementia who may be unable to skip a song or easily communicate distress, or users managing stress and anxiety, for whom arousal trajectories carry real clinical stakes~\cite{de2022music}. Offline learning is therefore not merely a data-efficiency choice; in this setting, it is an \emph{ethical constraint}: the policy must be trained and stress-tested without subjecting real listeners to unvalidated interventions on their affective state.

\paragraph{Production-policy exposure bias.} Songs in our logs were selected by a fixed production policy, so the data reflect that policy's coverage rather than free user choice. The policy combined preference-based retrieval (e.g., genre, rating-based similarity), rule-based playback constraints (e.g., skip management), and song-level affective tags annotated globally rather than per-listener. It scored each song in isolation, without considering listening-sequence effects or any individual user's measured affective response. Deploying an unvalidated new policy to collect counterfactual data for conventional off-policy correction would violate the ethical constraint above.

\paragraph{Our approach.} We build a \emph{rollout-based world model}~\cite{ha2018world}: a causal transformer trained on logged listening data that predicts all four feedback signals given a user history and a candidate next song. This model serves as an in-silico simulator, used to train and stress-test candidate policies entirely offline and to generate an unlimited supply of imagined rollouts from a finite logged dataset. The recommender policy is first initialized by behaviour cloning the production policy, then fine-tuned with Direct Preference Optimization (DPO)~\cite{rafailov2023direct} against a configurable multi-objective utility function over the simulator's predicted engagement, rating, valence, and arousal. DPO's KL penalty against the cloned reference acts as a safety constraint on policy drift. After training, the same world model stress-tests the chosen policy against counterfactual histories before any real-user exposure (Section~\ref{sec:stress}).

\paragraph{Contributions.} We frame this paper as an early validation, in a deployed setting, of a \emph{methodology} for affective music recommendation under ethical constraints. The specific contributions are: (i) a rollout-based world model that jointly predicts behavioural and affective feedback under a strict cold-start protocol; (ii) an architecture-and-embedding study comparing standard and \emph{factorized} causal transformers, with and without positional encoding, across two content-based song-embedding families (MERT and CLaMP~3); (iii) an offline preference-optimization pipeline that improves multi-objective predicted feedback while respecting diversity constraints; and (iv) deployment on LUCID's platform with a world-model-based pre-deployment safety workflow.

\section{Related Work}
\label{sec:related}

\paragraph{World models for recommendation.} Learning a generative model of user response and optimizing a policy against it has appeared in RecSim~\cite{ie2019recsim}, KuaiSim~\cite{zhao2023kuaisim}, and Spotify's playlist generation work~\cite{tomasi2023automatic}. The last of these shows that simulated feedback correlates with outcomes from real user studies, which directly motivates our use of a world model for offline evaluation. All of these systems model only behavioural signals. To our knowledge, none predict the session-level affective self-reports (valence, arousal) that are the optimization target of a functional music system. We adopt the \emph{rollout-based} world-model framework of \citet{ha2018world}, extended to behavioural and affective feedback jointly.

\paragraph{Offline RL and preference optimization for recommendation.} Offline RL has gained traction as a way to learn from logged data without running online experiments~\cite{gao2023alleviating,tomasi2023automatic}. The motivation in this previous work is typically data efficiency or engineering convenience; in our setting, offline learning is an ethical constraint on its own. Direct Preference Optimization~\cite{rafailov2023direct} replaces the separate reward model and the online RL step of Reinforcement Learning from Human Feedback (RLHF)~\cite{ouyang2022training} with a single supervised loss over preference pairs. Its application to sequential recommendation is still underexplored, and to our knowledge no prior work applies it to music recommendation with world-model-scored preference pairs. The KL term against the reference is particularly useful here because it prevents excessive drift from a cloned production policy whose clinical behaviour has been validated.

\paragraph{Multi-objective recommendation.} Multi-objective optimization in recommendation~\cite{jannach2023survey,stamenkovic2022choosing} conventionally treats accuracy, diversity, novelty, and business metrics as axes. Affective state rarely appears as a first-class objective; when it is considered, it is typically a contextual input or a post-hoc modifier. Our utility function admits predicted valence and arousal as first-class terms alongside engagement and rating; the specific weighting is chosen at DPO fine-tuning time (this paper uses an affect-only weighting).

\paragraph{Sequential and content-based music recommendation.} Transformer-based sequence models have become the dominant architectural choice for music recommendation~\cite{kang2018self,tran2024transformers}, and content-based song embeddings let a recommender absorb new catalogue additions without retraining~\cite{van2013deep,mei2025semantic}, a requirement for a platform whose catalogue grows continuously. We adopt both choices, representing songs via MERT~\cite{li2024mert} (a self-supervised acoustic model) and CLaMP~3~\cite{wu2025clamp} (a multimodal music-language model) embeddings rather than learned item identities.

\paragraph{Affective music recommendation.} \citet{hasan2025survey} survey affective recommender systems; recent work treats emotion as a dynamic state variable and learns policies that steer listeners toward desired states~\cite{jain2025causally}, and personalized music emotion recognition has emerged as a distinct subfield that models listener-specific affective responses rather than consensus labels~\cite{zhang2025personalized}. The closest prior recommender system is Moodify~\cite{de2022induced}, which uses Q-learning on data from 40 healthy adults in a lab setting. AMRS differs in three ways: it is trained fully offline on logged deployment data from clinical and consumer-wellness users; it explicitly corrects for production-policy exposure bias; and it admits a multi-objective utility function that can extend beyond emotion. Because Moodify is not publicly released, a head-to-head benchmark is not feasible.

\paragraph{Music-based intervention for affect and dementia care.} The dimensional valence-arousal representation~\cite{russell1980circumplex,eerola2012review} is standard in music affective research. \citet{russo2023developing} describe the clinical rationale for LUCID's digital therapeutic for dementia but do not present the recommendation architecture; we address that gap here. Meta-analyses of randomized trials~\cite{moreno2020music,lin2023effects} support music-based intervention for dementia care but identify the lack of standardized stimulus-selection as a recurring limitation that AMRS is designed to address.

\section{System Architecture}
\label{sec:architecture}

\begin{figure*}[!t]
\centering
\includegraphics[width=0.85\textwidth]{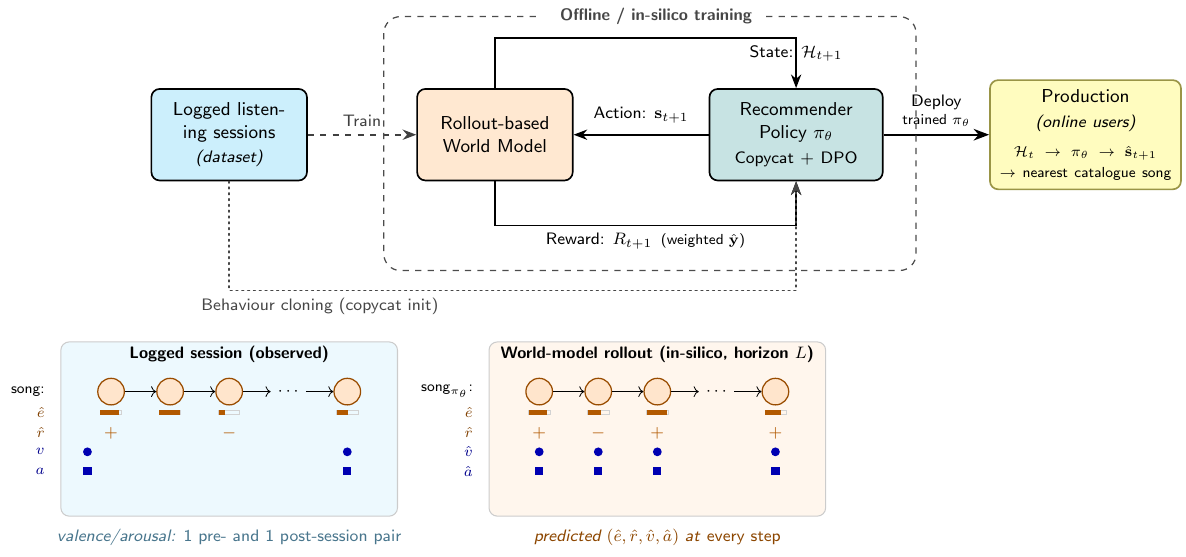}
\caption{AMRS architecture. Top: the offline training pipeline (dashed boundary); the trained policy is then deployed to production. Bottom: a logged session has engagement and (optional) ratings per song with valence-arousal observations only at session boundaries (post-session optional); a world-model rollout produces all four predictions $(\hat e, \hat r, \hat v, \hat a)$ at every step. This sparse-vs-dense affective contrast motivates the rollout-based world model.}
\Description{AMRS architecture diagram. Top: logged-sessions dataset, world model, policy, and production deployment; the world model and policy sit inside a dashed offline-training region. Bottom: two five-song panels contrasting sparse affective labels in a logged session with dense per-song predictions in a world-model rollout.}
\label{fig:framework}
\end{figure*}

AMRS has two components (Figure~\ref{fig:framework}). A rollout-based world model is trained on logged listening data to predict, at each step of a user history, the four feedback signals introduced above: engagement, binary rating, valence, and arousal. A recommender policy is then trained entirely against the world model: at each step it proposes a next song, the world model imputes the expected feedback, and a scalar utility score is computed. This separation provides \emph{safety} (no real-user exposure during training) and \emph{sample efficiency} (one trained world model generates unlimited simulated rollouts, even when real interaction data is scarce).

\subsection{Problem Formulation}
\label{sec:formulation}

We model the recommendation process as a Markov Decision Process (MDP) (Figure~\ref{fig:framework}). At time $t$, the state is the user's listening history $\mathcal{H}_t$, a sequence of song interactions, each carrying four signals: engagement rate $e\in[0,1]$ (fraction of the song listened to), binary rating $r$ (optional), and self-reported valence $v\in[0,1]$ and arousal $a\in[0,1]$, both measured before \emph{and} after each multi-song session (post-session values are optional). The recommender takes action $\mathbf{s}_{t+1}$ (the next song), the world model predicts $\hat{\mathbf{y}}_{t+1}=(\hat{e},\hat{r},\hat{v},\hat{a})_{t+1}$, and a reward is derived; the state transitions to $\mathcal{H}_{t+1}$ by appending the recommendation and its predicted feedback.

\subsection{User-History Representation}
\label{sec:history}

\paragraph{Song embeddings.} Songs are represented by content-based embeddings so the system can absorb catalogue additions without retraining, a requirement for a platform whose catalogue grows continuously. We compare two strategies: MERT~\cite{li2024mert} and CLaMP~3~\cite{wu2025clamp}. One-hot identity encodings are not used: they cannot generalize to new catalogue items and in pilot experiments performed substantially worse than either content-based alternative.

\paragraph{History token sequence.} Histories are represented as token sequences in which each token concatenates a song embedding, the available feedback signals, and binary masks indicating which signals are present for that token. This unified representation handles missing data naturally (valence and arousal are present only on session-boundary tokens; ratings may be absent at any step) and supports asking the model for any signal on any token that has a song attached. Interstitial ``session-boundary'' tokens separate listening sessions and carry the pre-session valence/arousal reports that are not tied to any individual song.

\subsection{World Model}
\label{sec:world_model}

The world model is a causal decoder-only transformer. We compare two variants in our experiments (a \emph{standard} transformer that right-shifts feedback to preserve causality, and a \emph{factorized} transformer that separates history encoding from feedback prediction) and describe only the deployed factorized variant in detail.

\paragraph{Factorized transformer.} A learnable Begin-Of-Sequence (BOS) token is prepended to the user history, and a causal transformer produces a sequence of user-history embeddings $\mathbf{z}_i$, each encoding the user history up to and including position $i$. Given a candidate next-song embedding $\mathbf{s}_{i+1}$, a lightweight MLP prediction head on the concatenation $[\mathbf{z}_i;\mathbf{s}_{i+1}]$ outputs the four-dimensional predicted feedback. Where positional information is used (the ``+PE'' rows of Table~\ref{tab:world_model}), rotary positional encodings~\cite{su2024roformer} are applied before the transformer layers; no-PE configurations use no explicit positional signal beyond the BOS token and causal mask. The factorization yields three practical advantages: (i) a dedicated history embedding useful for interpretability and the ablations of Section~\ref{sec:ablations}; (ii) parallel scoring of the full catalogue (one forward pass for the history, then a lightweight scoring of all candidate songs through the prediction head); and (iii) natural pairing between each song and its feedback in the input sequence, removing the right-shifting of feedback that the standard transformer requires to preserve causality.

\paragraph{Training.} Both architectures are trained with a composite loss: MSE on the continuous signals (engagement, valence, arousal) and BCE on the binary rating, applied only at positions where the corresponding target is observed. Pre-session valence and arousal enter as input context and contribute no loss; only post-session values are prediction targets. Training uses Monte Carlo Cross-Validation (MCCV) with 10 random 80/10/10 splits \emph{by user ID}, so validation and test users are strictly unseen during training (the ``strong generalization'' protocol of~\citet{marlin2009collaborative}). The same 10 splits are reused for every world-model variant we compare, enabling paired comparisons across configurations. Per-signal checkpoints are saved on validation loss, and the best per-signal checkpoint is loaded at evaluation. To vary the sequence contexts seen during training, each epoch re-samples 5 windows per user, each of random length drawn from $[50, 1000]$ tokens with a random starting point; validation and test use each user's full history.

\subsection{Recommender Policy}
\label{sec:recsys}

The recommender policy $\pi_\theta$ is a causal decoder-only transformer mapping a history $\mathcal{H}_t$ to a next-song embedding prediction $\hat{\mathbf{s}}_{t+1}$, matched to the catalogue via cosine similarity to produce the final recommendation.

\paragraph{Copycat initialization.} We first train a \emph{Copycat} model by behaviour cloning of the production policy. Given $\mathcal{H}_t$, the model predicts the embedding of the song that was actually played next; training minimizes an InfoNCE contrastive loss
{\small
\begin{equation}
\mathcal{L}_{\text{Copy}} = -\mathbb{E}\!\left[\log \frac{e^{\text{sim}(\hat{\mathbf{s}}, \mathbf{s}^+)/\tau}}{e^{\text{sim}(\hat{\mathbf{s}}, \mathbf{s}^+)/\tau} + \sum_{\mathbf{s}^-\in\mathcal{N}_t} e^{\text{sim}(\hat{\mathbf{s}}, \mathbf{s}^-)/\tau}}\right]
\label{eq:copycat}
\end{equation}}%
where $\hat{\mathbf{s}}=f(\mathcal{H}_t)$, $\mathbf{s}^+$ is the true next song, $\mathcal{N}_t$ is a set of unplayed song embeddings, $\text{sim}$ is cosine similarity, and $\tau$ is a temperature. The production policy combined preference-based retrieval, rule-based playback constraints, and global per-song affective tags, operating on each song in isolation rather than over a listening sequence. The tags target affect at the population level rather than against measured individual response, so Copycat imitation inherits a tag-based behavioural prior on which DPO can improve via measured-feedback optimization.

\paragraph{Direct preference optimization.} Starting from the Copycat reference $\pi_{\text{ref}}$, we fine-tune with DPO~\cite{rafailov2023direct}. For each state $x=\mathcal{H}_t$, we construct preference pairs $(y_w,y_l)$ where $y_w$ has a higher predicted utility score (preferred song) and $y_l$ is the less preferred song under the world model:
{\small
\begin{equation}
\mathcal{L}_{\text{DPO}} = -\mathbb{E}_{(x,y_w,y_l)}\!\left[\log\sigma\!\left(\beta\log\tfrac{\pi_\theta(y_w|x)}{\pi_{\text{ref}}(y_w|x)} - \beta\log\tfrac{\pi_\theta(y_l|x)}{\pi_{\text{ref}}(y_l|x)}\right)\right]
\label{eq:dpo}
\end{equation}}%
where $\beta$ controls the KL penalty against the reference.

\paragraph{Preference-pair construction.} We use \emph{negative history sampling}: for each time step where the production recommendation received negative feedback, we identify a catalogue replacement that would likely receive positive feedback under the world model. The candidate pool of size $M$ combines \emph{corrective sampling} (top-$M/2$ nearest neighbours of the poorly-received song in embedding space) with \emph{exploratory sampling} ($M/2$ uniform draws from the rest of the catalogue). The candidate with the highest utility score becomes $y_w$; the original poorly-received recommendation becomes $y_l$. The construction is explicitly counterfactual: it targets the production policy's mistakes and proposes world-model-scored corrections.

\paragraph{Rollout seeding.} Recommender training and evaluation use a \emph{window strategy}: each rollout's initial history $\mathcal{H}_0$ is a contiguous segment (a \emph{window}) of a user's logged sequence, with 5 windows per user, each of random length drawn from $[10, 100]$ tokens with a random starting point. This multiplies the effective training signal by $\sim$5$\times$ and gives denser coverage of each user's history; it forces the policy to act on partial histories, matching both the cold-start regime evaluated by the strong-generalization protocol and the partial-history conditioning expected at deployment. Training, fine-tuning, and evaluation use the same windows so cross-policy comparisons are reproducible.

\paragraph{Multi-seed training and evaluation.} All four policies under comparison (Copycat, DPO, and the Random/Greedy baselines defined in Section~\ref{sec:recommender_results}) are run with $N{=}5$ random seeds, shared across policies so cross-policy comparisons are paired. Reported metrics are cross-seed mean$\,\pm\,$standard deviation (SD).

\subsection{Multi-Objective Utility Function and Safety Constraints}
\label{sec:utility}

The utility function maps predicted feedback at each step to a scalar utility $R_t$ (equal to the reward in the reinforcement learning framing), a weighted combination of the four predicted signals:
{\small
\begin{equation}
R_t = \lambda_r\hat{r}_t + \lambda_e\hat{e}_t + \lambda_v\hat{v}_t + \lambda_a\hat{a}_t,\ \lambda_\bullet\in[-1,1]
\label{eq:utility}
\end{equation}}%
DPO scores preference pairs on this per-step utility (Section~\ref{sec:recsys}); evaluation reports the per-signal rollout averages $\tfrac{1}{L}\sum_t \hat{y}_t$ that compose this utility. The specific weighting is set when DPO is fine-tuned and depends on the deployment's product or clinical objective.

Safety in AMRS is encoded through three mechanisms, each tied to a specific ethical concern. (i)~\emph{Fully offline training and stress-testing.} All policy learning occurs against the world model; the same model is also used post-training to probe the policy against counterfactual histories before any real-user exposure (Section~\ref{sec:stress}). (ii)~\emph{Exposure-bias correction via Copycat and KL.} The logged data reflects the production policy's coverage rather than free user choice; we anchor the new policy by cloning the production policy and applying a KL penalty against it in DPO, monitored by an explicit policy-adherence evaluation (Section~\ref{sec:policy_alignment}). (iii)~\emph{Distributional robustness.} Diversity metrics (coverage, normalized entropy, intra-list diversity (ILD), Gini) are tracked jointly with predicted feedback to detect recommendation collapse.

\section{Evaluation}
\label{sec:evaluation}

We evaluate AMRS along three axes: predicted feedback quality (rollout performance across all four signals), distributional robustness (coverage, entropy, ILD, Gini), and policy alignment (adherence to the logged production policy).

\subsection{Dataset}
\label{sec:dataset}

Experiments use a proprietary dataset drawn from LUCID's deployed platform, pooling logs from the clinical and consumer-wellness applications. In both, each listening session is a curated sequence of songs selected by the platform's fixed production policy rather than by free user choice, so observed sequences reflect that policy's coverage. Implicit signals such as ``played vs.\ not played'' are therefore imperfect preference proxies, a universal caveat in offline recommender evaluation. This matters especially here because the candidate-replacement step in DPO depends on the logged interactions. Table~\ref{tab:datasets} summarizes the dataset. Data is split 80/10/10 by user ID using MCCV with 10 random splits, so all interactions from a single user remain within a single split (strong generalization / cold-start evaluation).

\begin{table}[!t]
\caption{Dataset summary. Pooled logs from LUCID's deployed clinical and consumer-wellness music applications.}
\Description{Dataset summary: 939 users, 8784 sessions, 57822 user-song interactions, 5992 unique songs from a 6988-song catalogue. 16927 positive and 2561 negative ratings. Pre-session affect labels for all sessions; post-session for 5319 (about 60 percent).}
\label{tab:datasets}
\centering
\small
\begin{tabular}{@{}lr@{}}
\toprule
\textbf{Statistic} & \textbf{Count} \\
\midrule
Unique users                & 939   \\
Unique sessions             & 8{,}784 \\
User-song interactions      & 57{,}822 \\
Unique songs played         & 5{,}992 \\
Song catalogue size           & 6{,}988 \\
Positive ratings            & 16{,}927 \\
Negative ratings            & 2{,}561 \\
Pre-session valence/arousal & 8{,}784 \\
Post-session valence/arousal & 5{,}319 \\
\bottomrule
\end{tabular}
\end{table}

\subsection{Evaluation Metrics}
\label{sec:metrics}

\paragraph{World-model metrics.} For continuous signals (engagement, valence, arousal) we report $R^2$, the fraction of variance explained relative to a mean predictor; negative values indicate worse-than-mean performance. For the binary rating signal we report ROC AUC, computed over tokens where a rating was actually observed (Table~\ref{tab:datasets}).

\paragraph{Recommender metrics.} Predicted feedback quality is evaluated via rollouts under the world model: for each seed, the recommender generates $L=5$ songs and we report mean predicted engagement, rating, valence, and arousal. Diversity uses four complementary metrics: \emph{coverage} (fraction of catalogue items appearing at least once across rollouts), \emph{normalized entropy} (entropy of the recommendation distribution divided by the catalogue entropy), \emph{ILD} (mean pairwise cosine distance between content embeddings within a rollout, normalized by the catalogue mean), and the \emph{Gini coefficient} (inequality of the item-occurrence distribution). Coverage and normalized entropy address breadth across the catalogue; ILD addresses within-rollout similarity; Gini complements normalized entropy by quantifying long-tail inequality. A collapsed policy shows near-zero coverage and entropy, near-zero ILD, and Gini near 1, which is exactly the pattern produced by greedy optimization (Table~\ref{tab:feedback_window}).

\paragraph{Policy-alignment metrics.} We assess adherence to the logged production policy at three levels of strictness, reporting Hit@10, mean reciprocal rank (MRR), normalized discounted cumulative gain (NDCG@10), and (at Level~3 only) accuracy. This is a diagnostic, not a target: a policy that perfectly imitates the production policy cannot improve on it, and we expect DPO to show \emph{controlled} drift from the cloned baseline. \emph{Level~2 (user-history imitation)} counts an item as relevant if it appears anywhere in that user's logged history (the primary level of analysis); \emph{Level~3 (time-step imitation)} counts an item as relevant only if it matches the exact song at the corresponding time step (a strictness upper bound that no learned policy is expected to reach under cold start). We also considered \emph{Level~1 (catalogue exposure)}, where an item is relevant if it appears in any user's logged history, but exclude it because relevance defined on exposure can be artificially inflated by production-policy sampling bias.

\subsection{World-Model Results}
\label{sec:results}

Table~\ref{tab:world_model} reports world-model results across architecture and embedding configurations, separating song-level behavioural signals (engagement, rating) from session-level affective signals (valence, arousal). The shaded row is the configuration selected by average rank across signals with equal weights, using a primary metric per signal (MSE for continuous targets, AUC for the binary rating).

\begin{table*}[!t]
\caption{World-model results across architecture and embedding configurations. Mean (SD) over 10 MCCV test splits. PE = rotary positional encoding~\cite{su2024roformer}. \textbf{Bold} = best per-metric mean; shaded row = overall-best configuration by equal-weight average rank. Stars: paired $t$-test vs.\ shaded row ($^{*}p<0.05$, $^{**}p<0.01$, $^{***}p<0.001$).}
\Description{Eight-row table comparing 2 transformer architectures across 4 embedding/PE configurations on engagement, rating, valence, and arousal. The factorized MERT (no-PE) row is shaded as overall-best by average rank; per-metric peaks are bolded. Patterns are discussed in the main text.}
\label{tab:world_model}
\centering
\small
\begin{tabular}{@{}ll cc | c | cc | cc @{}}
\toprule
& & \multicolumn{2}{c|}{\textbf{Engagement}} & \textbf{Rating} & \multicolumn{2}{c|}{\textbf{Valence}} & \multicolumn{2}{c}{\textbf{Arousal}} \\
\textbf{Arch.} & \textbf{Config} & MSE$\downarrow$ & $R^2\!\uparrow$\,(\%) & AUC$\uparrow$\,(\%) & MSE$\downarrow$ & $R^2\!\uparrow$\,(\%) & MSE$\downarrow$ & $R^2\!\uparrow$\,(\%) \\
\midrule
\multirow{4}{*}{Transformer}
 & CLaMP~3      & 0.059\,{\scriptsize(0.015)} & 10.3\,{\scriptsize(10.0)}$^{*}$ & 68.8\,{\scriptsize(8.4)}$^{**}$ & 0.029\,{\scriptsize(0.005)}$^{**}$ & 36.7\,{\scriptsize(10.1)}$^{**}$ & 0.040\,{\scriptsize(0.008)} & 32.5\,{\scriptsize(15.2)} \\
 & MERT         & \textbf{0.055\,{\scriptsize(0.016)}} & \textbf{16.9\,{\scriptsize(7.0)}} & 72.9\,{\scriptsize(7.4)}$^{*}$ & \textbf{0.026\,{\scriptsize(0.005)}} & \textbf{43.3\,{\scriptsize(11.1)}} & \textbf{0.036\,{\scriptsize(0.010)}} & \textbf{40.2\,{\scriptsize(16.2)}} \\
 & CLaMP~3 + PE & 0.058\,{\scriptsize(0.018)} & 12.0\,{\scriptsize(6.4)}$^{*}$ & 64.3\,{\scriptsize(8.8)}$^{***}$ & 0.031\,{\scriptsize(0.007)} & 31.6\,{\scriptsize(20.0)} & 0.038\,{\scriptsize(0.008)} & 35.9\,{\scriptsize(13.6)} \\
 & MERT + PE    & 0.057\,{\scriptsize(0.017)} & 13.6\,{\scriptsize(5.5)} & 67.7\,{\scriptsize(7.3)}$^{**}$ & 0.031\,{\scriptsize(0.006)}$^{***}$ & 33.6\,{\scriptsize(11.8)}$^{***}$ & 0.038\,{\scriptsize(0.010)} & 35.7\,{\scriptsize(17.6)} \\
\midrule
\multirow{4}{*}{Factorized}
 & CLaMP~3      & 0.056\,{\scriptsize(0.017)} & 15.6\,{\scriptsize(5.7)} & 71.7\,{\scriptsize(7.5)} & 0.029\,{\scriptsize(0.006)} & 39.0\,{\scriptsize(11.2)} & \textbf{0.036\,{\scriptsize(0.008)}} & 39.6\,{\scriptsize(14.2)} \\
 & \cellcolor{gray!15}MERT
       & \cellcolor{gray!15}\textbf{0.055\,{\scriptsize(0.014)}}
       & \cellcolor{gray!15}16.4\,{\scriptsize(7.1)}
       & \cellcolor{gray!15}\textbf{74.1\,{\scriptsize(7.8)}}
       & \cellcolor{gray!15}0.027\,{\scriptsize(0.005)}
       & \cellcolor{gray!15}42.6\,{\scriptsize(10.3)}
       & \cellcolor{gray!15}0.037\,{\scriptsize(0.012)}
       & \cellcolor{gray!15}37.9\,{\scriptsize(18.4)} \\
 & CLaMP~3 + PE & 0.057\,{\scriptsize(0.018)} & 14.2\,{\scriptsize(7.1)} & 73.1\,{\scriptsize(7.8)} & 0.030\,{\scriptsize(0.005)}$^{**}$ & 36.3\,{\scriptsize(9.0)}$^{**}$ & 0.039\,{\scriptsize(0.010)} & 35.1\,{\scriptsize(18.0)} \\
 & MERT + PE    & \textbf{0.055\,{\scriptsize(0.018)}} & 16.2\,{\scriptsize(6.3)} & 73.4\,{\scriptsize(7.3)} & 0.029\,{\scriptsize(0.004)} & 37.4\,{\scriptsize(11.0)}$^{*}$ & 0.039\,{\scriptsize(0.012)} & 34.5\,{\scriptsize(19.2)} \\
\bottomrule
\end{tabular}
\end{table*}

Three patterns are worth highlighting, all of which we expect to generalize beyond the specific world-model instance.

\textit{(i) MERT without positional encoding is the most robust configuration across signals.} The factorized MERT (no-PE) row is overall-best by average rank across signals and ranks near the top on every individual metric. The transformer MERT (no-PE) variant matches it on per-metric peaks for both engagement and the affective signals. We select the factorized variant for deployment because it produces the dedicated history embedding used by the ablations and the recommender, and because its parallel scoring path is computationally cheaper at catalogue scale (Section~\ref{sec:world_model}). No-PE configurations generally match or outperform their +PE counterparts, with the strongest effect under MERT embeddings, suggesting that the factorized model's BOS token and causal mask already encode sequence position implicitly and make explicit positional encoding redundant or, in several cases, slightly harmful.

\textit{(ii) MERT outperforms CLaMP~3 on most signals.} Holding architecture and PE fixed, MERT matches or outperforms CLaMP~3 on engagement, rating, and valence in every configuration. On arousal, the picture is mixed: CLaMP~3 has a small edge in the factorized configurations and Transformer + PE, while MERT wins only in the Transformer no-PE setting. We read this as MERT's acoustic representation being the more useful prior for per-token feedback prediction, even though CLaMP~3 was designed for higher-level musical semantics; fine-tuned or task-aligned embeddings are a direction for future work.

\textit{(iii) Session-level affect is predictable under a strict cold-start protocol.} Valence $R^2$ reaches 43.3\% and arousal $R^2$ reaches 40.2\% under the best per-metric configurations, with the deployed factorized MERT row at 42.6\% and 37.9\%. These are substantial given that validation and test users are strictly unseen during training and that session-level labels are an order of magnitude sparser than song-level interactions. Reported music emotion recognition (MER) benchmarks span a wide range: studies on DEAM, PMEmo, and EmoMusic datasets, where each clip carries a single consensus label, report $R^2$ of roughly 50-65\% for valence and 60-80\% for arousal~\cite{kang2025towards}. Our 42.6\% and 37.9\% are below this range, but the task is harder: strict cold-start, and per-session listener-specific affect rather than per-stimulus consensus, a setting known to degrade performance~\cite{zhang2025personalized,eerola2012review}. The recommender depends on ranking candidate songs by predicted affect rather than absolute calibration, and these $R^2$ values support that.

\subsection{Ablations}
\label{sec:ablations}

Table~\ref{tab:ablation} reports a leave-one-feature-out ablation on the selected factorized MERT (no-PE) configuration, removing one input component at a time during training \emph{and} test.

\begin{table*}[!t]
\caption{Leave-one-feature-out ablation on the factorized MERT (no-PE) world-model base. Each row removes one input during both training and test. Mean (SD) over 10 MCCV test splits. \textbf{Bold} = best per-metric mean. Stars: paired $t$-test vs.\ base ($^{*}p<0.05$, $^{**}p<0.01$, $^{***}p<0.001$).}
\Description{Leave-one-feature-out ablation on the factorized MERT (no-PE) base. Removing user history or feedback signals significantly degrades every metric (p<0.01 or better); other ablations are within noise. Detailed patterns are discussed in the main text.}
\label{tab:ablation}
\centering
\small
\begin{tabular}{@{}l cc | c | cc | cc @{}}
\toprule
& \multicolumn{2}{c|}{\textbf{Engagement}} & \textbf{Rating} & \multicolumn{2}{c|}{\textbf{Valence}} & \multicolumn{2}{c}{\textbf{Arousal}} \\
\textbf{Configuration} & MSE$\downarrow$ & $R^2\!\uparrow$\,(\%) & AUC$\uparrow$\,(\%) & MSE$\downarrow$ & $R^2\!\uparrow$\,(\%) & MSE$\downarrow$ & $R^2\!\uparrow$\,(\%) \\
\midrule
Factorized MERT no-PE (base)
 & 0.055\,{\scriptsize(0.014)} & 16.4\,{\scriptsize(7.1)} & \textbf{74.1\,{\scriptsize(7.8)}}
 & 0.027\,{\scriptsize(0.005)} & 42.6\,{\scriptsize(10.3)} & 0.037\,{\scriptsize(0.012)} & 37.9\,{\scriptsize(18.4)} \\
\midrule
$-$ User history        & 0.067\,{\scriptsize(0.021)}$^{***}$ & $-$1.4\,{\scriptsize(3.6)}$^{***}$ & 56.9\,{\scriptsize(4.9)}$^{***}$ & 0.045\,{\scriptsize(0.007)}$^{***}$ & 4.2\,{\scriptsize(4.8)}$^{***}$ & 0.057\,{\scriptsize(0.007)}$^{***}$ & 4.6\,{\scriptsize(3.3)}$^{***}$ \\
$-$ Feedback signals    & 0.062\,{\scriptsize(0.020)}$^{**}$ & 5.4\,{\scriptsize(5.5)}$^{***}$ & 59.3\,{\scriptsize(6.6)}$^{***}$ & 0.045\,{\scriptsize(0.010)}$^{***}$ & 3.2\,{\scriptsize(15.3)}$^{***}$ & 0.061\,{\scriptsize(0.014)}$^{***}$ & $-$0.5\,{\scriptsize(18.1)}$^{***}$ \\
$-$ Historical songs    & 0.054\,{\scriptsize(0.014)} & \textbf{18.1\,{\scriptsize(4.7)}} & 74.0\,{\scriptsize(7.1)} & 0.026\,{\scriptsize(0.005)} & 44.1\,{\scriptsize(8.1)} & 0.035\,{\scriptsize(0.012)} & 41.2\,{\scriptsize(20.9)} \\
$-$ Max lookback 10     & \textbf{0.053\,{\scriptsize(0.013)}} & 17.9\,{\scriptsize(6.1)} & 73.5\,{\scriptsize(6.7)} & \textbf{0.025\,{\scriptsize(0.005)}} & \textbf{45.8\,{\scriptsize(5.3)}} & \textbf{0.034\,{\scriptsize(0.007)}} & \textbf{42.7\,{\scriptsize(12.3)}} \\
$-$ Max lookback 5      & 0.054\,{\scriptsize(0.014)} & 16.9\,{\scriptsize(6.5)} & 73.5\,{\scriptsize(7.1)} & 0.028\,{\scriptsize(0.002)} & 39.0\,{\scriptsize(7.6)} & 0.036\,{\scriptsize(0.007)} & 38.9\,{\scriptsize(11.2)} \\
$-$ Recommended song    & 0.055\,{\scriptsize(0.016)} & 16.2\,{\scriptsize(6.2)} & 73.4\,{\scriptsize(7.9)} & 0.028\,{\scriptsize(0.006)} & 40.0\,{\scriptsize(12.3)} & 0.037\,{\scriptsize(0.010)} & 37.9\,{\scriptsize(17.6)} \\
\bottomrule
\end{tabular}
\end{table*}

Removing user history or all feedback signals significantly degrades every signal ($p<0.01$ or better in all cases, with most at $p<0.001$): historical behaviour and past feedback together carry the dominant predictive signal at this data scale. The remaining ablations are within noise. Max-lookback~10 is the best ablation by average rank, with the largest nominal gains on session-level affect (valence $R^2$ 45.8\%, arousal $R^2$ 42.7\%), suggesting that recent context captures most of the predictive signal. Removing historical songs while keeping feedback and session-boundary tokens yields small gains across signals (engagement, valence, and arousal), and removing the recommended song produces small drops across signals. Together, these suggest that user behavioural signals (engagement, rating, valence/arousal) carry the dominant predictive weight, with song-content embeddings playing a secondary role and most of that role concentrated in the most recent history. The base is retained as the deployed configuration since these movements sit within noise.

\paragraph{Centrality of the world model.} The two collapsing ablations ($-$user history, $-$feedback signals) identify the inputs the world model must have to produce useful predictions (input sufficiency). When the world model is uninformative, every downstream component (DPO preference scoring, the rollout-based utility function in Equation~\ref{eq:utility}, and stress-testing in Section~\ref{sec:stress}) loses its signal. The ablations therefore support positioning the rollout-based world model as AMRS's central component rather than an auxiliary one.

\subsection{Recommender Results}
\label{sec:recommender_results}

Recommender results come from rollout simulations under the same world model used to construct DPO preference pairs. Because DPO is both optimized and evaluated against this model, gains may partly reflect simulator exploitation rather than genuine user-outcome improvement; predictions may regress toward the training-data mean and compounding errors may distort long-horizon estimates. Unlike the ablation findings, which depend only on input sufficiency, the recommender results below are conditional on this specific world model.

\paragraph{Optimization target and role of song-level metrics.} The DPO policy is fine-tuned on an \emph{affect-only} utility ($\lambda_v=\lambda_a=0.5,\ \lambda_e=\lambda_r=0$): valence and arousal are the optimization target. Rating and engagement are held out of the objective as a no-degradation check; a small drop on them is expected and acceptable, leaving them available for future utility weights.

We compare four policies. \emph{Random} samples uniformly from the catalogue (broad coverage, mean-reverting feedback; a lower bound). \emph{Greedy} scores every candidate under the world model and selects the one with highest predicted feedback (severe distributional collapse; a degenerate upper bound). \emph{Copycat} is the behaviour-cloned model trained to imitate the production policy. \emph{DPO} is the Copycat fine-tuned against the affect-only utility function described above.

Table~\ref{tab:feedback_window} reports predicted feedback and diversity for all four policies under the window rollout-seeding strategy (Section~\ref{sec:recsys}), averaged across $N{=}5$ training seeds. Song-level signals (rating, engagement) are held out of the DPO objective and shown as a no-degradation check, visually grouped to keep that distinction clear.

\begin{table*}[!t]
\caption{Predicted feedback and diversity across recommendation policies
(window strategy, test split, $L{=}5$ songs, $N{=}5$ seeds). Rating and
engagement are held out of the DPO objective and shown as a no-degradation
check. Cross-seed mean (SD). \textbf{Bold} = best among learned
policies (Copycat, DPO). Stars: paired $t$-test vs.\ DPO ($^{*}p<0.05$,
$^{**}p<0.01$, $^{***}p<0.001$).}
\Description{Predicted affect (DPO objective), predicted behavioural feedback held out of the objective, and diversity for Copycat, DPO, Random, and Greedy under the window strategy across 5 seeds. Vertical rules separate the three column groups. DPO improves over Copycat on the affect objective; the held-out rating and engagement drop only slightly, confirming song-level prediction capability is preserved. Greedy collapses diversity; Random has broad coverage but low affect.}
\label{tab:feedback_window}
\centering
\small
\begin{tabular}{@{}l cc | cc | cccc@{}}
\toprule
& \multicolumn{2}{c|}{\textbf{Affect (DPO objective)}} & \multicolumn{2}{c|}{\textbf{Behavioural (held out)}} & \multicolumn{4}{c}{\textbf{Diversity}} \\
\cmidrule(lr){2-3} \cmidrule(lr){4-5} \cmidrule(lr){6-9}
\textbf{Model} & Valence$\uparrow$ & Arousal$\uparrow$ & Rating$\uparrow$ & Engagement$\uparrow$ & Cov.$\uparrow$ & Norm.\ Ent.$\uparrow$ & ILD$\uparrow$ & Gini$\downarrow$ \\
\midrule
Copycat & 0.480\,{\scriptsize(0.009)}$^{***}$ & 0.433\,{\scriptsize(0.011)}$^{***}$ & \textbf{0.866\,{\scriptsize(0.025)}}$^{***}$ & \textbf{0.765\,{\scriptsize(0.035)}}$^{***}$ & 0.029\,{\scriptsize(0.003)}$^{**}$ & 0.450\,{\scriptsize(0.012)}$^{**}$ & 2.166\,{\scriptsize(0.091)}$^{***}$ & 0.993\,{\scriptsize(0.001)}$^{*}$ \\
\textbf{DPO} & \textbf{0.499\,{\scriptsize(0.012)}} & \textbf{0.449\,{\scriptsize(0.012)}} & 0.830\,{\scriptsize(0.028)} & 0.725\,{\scriptsize(0.032)} & 0.022\,{\scriptsize(0.002)} & 0.478\,{\scriptsize(0.011)} & 1.763\,{\scriptsize(0.091)} & 0.993\,{\scriptsize(0.001)} \\
\midrule
\multicolumn{9}{@{}l}{\textit{Baselines:}} \\[2pt]
Random & 0.202\,{\scriptsize(0.010)}$^{***}$ & 0.171\,{\scriptsize(0.008)}$^{***}$ & 0.876\,{\scriptsize(0.010)}$^{**}$ & 0.643\,{\scriptsize(0.010)}$^{**}$ & 0.287\,{\scriptsize(0.003)}$^{***}$ & 0.853\,{\scriptsize(0.001)}$^{***}$ & 0.997\,{\scriptsize(0.008)}$^{***}$ & 0.750\,{\scriptsize(0.004)}$^{***}$ \\
Greedy & 0.610\,{\scriptsize(0.012)}$^{***}$ & 0.550\,{\scriptsize(0.011)}$^{***}$ & 0.782\,{\scriptsize(0.023)}$^{***}$ & 0.572\,{\scriptsize(0.032)}$^{***}$ & 0.002\,{\scriptsize(0.000)}$^{***}$ & 0.221\,{\scriptsize(0.007)}$^{***}$ & 0.480\,{\scriptsize(0.027)}$^{***}$ & 0.999\,{\scriptsize(0.000)}$^{***}$ \\
\bottomrule
\end{tabular}
\end{table*}

The pattern matches the offline preference-optimization story. On the DPO objective, valence and arousal improve over Copycat by 4.0\% and 3.7\% relative (both $p<0.001$ paired across seeds), with the gaps roughly 2$\times$ and 1.5$\times$ the per-policy seed-SD respectively. On the held-out song-level signals (middle group of Table~\ref{tab:feedback_window}), rating and engagement decline by 4.2\% and 5.2\%, at roughly the same scale as the affect improvement; these small drops indicate that DPO didn't sacrifice song-level signals to optimize affect (unlike Greedy), leaving rating and engagement available for future utility weights with $\lambda_e,\lambda_r > 0$. DPO's diversity profile shifts modestly from Copycat, with all four shifts statistically significant ($p<0.05$ or better, paired across seeds): coverage and ILD drop ($0.029{\to}0.022$, $2.17{\to}1.76$) while normalized entropy rises ($0.450{\to}0.478$) and Gini remains the same. This indicates that DPO concentrates on a smaller catalogue slice while distributing its picks within that slice more evenly than Copycat. DPO stays well clear of the distributional collapse Greedy exhibits (coverage $0.002$, ILD $0.480$). Among baselines, Greedy maximises predicted affect at the cost of near-zero coverage and the lowest ILD in the table; Random has the broadest coverage but the lowest predicted affect, reflecting the world model's mean-reverting predictions for unseen song-history combinations.

We caution that the absolute magnitude of DPO's affect gain over Copycat is conditional on the world model used to score these rollouts. The qualitative pattern, in which DPO improves on its objective, declines marginally on the held-out signals, and avoids collapse, is the more transferable claim.

\subsection{Policy Alignment}
\label{sec:policy_alignment}

Table~\ref{tab:adherence} reports user-wise (Level~2) and time-step (Level~3) adherence to policy under the window strategy. Adherence is a diagnostic, not a target: a policy that only imitates the production policy cannot improve on it, so we expect Copycat to dominate the table by construction and DPO to show controlled drift away from it.

\begin{table}[!t]
\caption{Policy adherence under the window strategy ($N{=}5$ seeds). \emph{Level~2} (user-wise): an item is relevant if it appears in the user's logged history. \emph{Level~3} (time-step): an item is relevant only if it matches the exact song at the same time step. Values are cross-seed means. Cross-seed SDs are $\leq 0.005$ at Level~3; at Level~2, $\leq 0.05$ on train/test and $\leq 0.08$ on validation. Highest value per row in bold.}
\Description{Policy adherence (Level 2 user-wise and Level 3 time-step) for Copycat, DPO, Random, and Greedy across 5 seeds and train/val/test splits. Copycat dominates by construction; DPO sits between Copycat and the baselines, indicating controlled drift.}
\label{tab:adherence}
\centering
\small
\begin{tabular}{@{}llcccc@{}}
\toprule
\textbf{Split} & \textbf{Metric} & \textbf{Copycat} & \textbf{DPO} & \textbf{Random} & \textbf{Greedy} \\
\midrule
\multicolumn{6}{@{}l}{\textit{Level~2 (user-wise)}} \\
Train & Hit@10  & \textbf{0.937} & 0.674 & 0.039 & 0.163 \\
 & MRR     & \textbf{0.732} & 0.393 & 0.018 & 0.052 \\
 & NDCG@10 & \textbf{0.359} & 0.179 & 0.015 & 0.062 \\
Val  & Hit@10  & \textbf{0.266} & 0.206 & 0.037 & 0.100 \\
 & MRR     & \textbf{0.114} & 0.092 & 0.013 & 0.029 \\
 & NDCG@10 & \textbf{0.123} & 0.101 & 0.017 & 0.043 \\
Test & Hit@10  & \textbf{0.279} & 0.186 & 0.036 & 0.084 \\
 & MRR     & \textbf{0.121} & 0.083 & 0.012 & 0.038 \\
 & NDCG@10 & \textbf{0.130} & 0.089 & 0.016 & 0.048 \\
\midrule
\multicolumn{6}{@{}l}{\textit{Level~3 (time-step, test only)}} \\
Test & Accuracy & \textbf{0.005} & 0.003 & 0.000 & 0.000 \\
 & Hit@10  & \textbf{0.033} & 0.016 & 0.002 & 0.002 \\
 & MRR     & \textbf{0.012} & 0.006 & 0.000 & 0.001 \\
 & NDCG@10 & \textbf{0.015} & 0.007 & 0.001 & 0.001 \\
\bottomrule
\end{tabular}
\end{table}

The adherence pattern matches expectations. At Level~2, Copycat dominates the training split (Hit@10$=0.94$, MRR$=0.73$ across seeds), confirming successful behaviour cloning; on unseen validation and test users this drops sharply (test Hit@10$=0.28$), the expected cold-start behaviour for a strong-generalization protocol. DPO sits below Copycat at every split (test Hit@10 $0.19$ vs.\ $0.28$) but well above Random and Greedy: the KL penalty against the cloned reference has produced controlled drift toward the affect objective rather than complete replacement of the production policy. At test, every inter-policy gap (Copy-DPO $\approx 0.09$, DPO-Greedy $\approx 0.10$, DPO-Random $\approx 0.15$ on Hit@10) exceeds the worst-case cross-seed SD ($\leq 0.05$) by roughly 2$\times$ or more, so the ordering is robust to seed variability. At Level~3, every learned policy is far below 4\% Hit@10 on the test split, confirming that exact time-step matching is not an attainable target under cold start and that Level~2 is the appropriate level for substantive comparison. Greedy's non-zero Level~2 numbers reflect its concentration on a small set of popular songs that appear in many users' histories, not genuine imitation.

\section{Discussion and Deployment}
\label{sec:discussion}

\subsection{The World Model as a Stress-Testing Tool}
\label{sec:stress}

The trained world model also stress-tests candidate policies before deployment. Conditioning on synthetic or perturbed histories (atypical pre-session valence/arousal, cold-start users) surfaces undesirable behaviours such as coverage collapse and large arousal swings, which can be converted into deployment-time guardrails (e.g., maximum per-step arousal delta, minimum coverage floor) before any real-user exposure.

\subsection{Limitations and Generalization}

The dataset is single-platform, collected under a single production policy, so cross-platform validation remains future work. The absolute magnitudes of DPO's gains and the per-metric optima are also specific to this trained world model and dataset; the qualitative patterns reported above (Greedy collapses, DPO drifts in a controlled way from Copycat, the ablation input-sufficiency conditions) are architectural, consistent with the offline-RL literature, and are the more transferable claims.

\subsection{Ethical Considerations}

In this setting, offline-only training is an ethical imperative: deploying untested policies that influence affective state amounts to uncontrolled experimentation, and clinical users may have limited ability to communicate dissatisfaction. Our safeguards (no real-user exposure during training, Copycat anchoring, KL-penalised DPO, diversity and policy-alignment monitoring) are necessary, not sufficient: online evaluation under clinical oversight remains the ultimate validation step.

\subsection{Future Work}

On the recommender side: exploring different utility weights, including dynamic switching based on user-stated intention (energize vs.\ calm); extending DPO from single-step pairs to multi-song sub-sequences for sequential therapeutic structure (e.g., arousal-reduction toward sleep); and a lightweight re-ranker for production constraints such as repetition limits and coverage floors. On the world model: richer user context (probabilistic models, task-aligned embeddings, time-of-day, biometric signals), stronger acoustic and semantic embeddings, longer context windows, and continuous retraining as new logged sessions accumulate to keep the simulator aligned with platform drift. Finally, carefully scoped online evaluation under clinical oversight is the remaining validation step.

\section{Conclusion}
\label{sec:conclusion}

We present AMRS, deployed on LUCID's music platform, in which a rollout-based world model serves as both an offline simulator and a pre-deployment stress-testing tool. The world model predicts behavioural and affective feedback with usable fidelity under cold start, and the DPO policy improves predicted valence and arousal over the cloned baseline while preserving song-level prediction capability and avoiding distributional collapse. Ablations establish the centrality of the world model: removing its inputs degrades the simulator to near-random, leaving DPO scoring, the rollout-based utility function, and stress-testing without signal. The joint behavioural and affective world model is therefore the central technical contribution; the recommender pipeline and safety workflow follow from it.

\begin{acks}
We thank our collaborators at Mila --- Qu\'{e}bec AI Institute for the research partnership; the LUCID product, engineering, and science teams for building the applications and grounding the research; and the consumer-wellness listeners, clinical participants, and care teams at our partner facilities, whose anonymized listening data made this work possible. We'd also like to thank the governments of Canada and Qu\'{e}bec for various granting programs that have made this research possible. 
\end{acks}

\bibliographystyle{ACM-Reference-Format}
\bibliography{references}

\end{document}